\documentclass{article}
\usepackage{spconf,amsmath,graphicx}
\usepackage{amsmath}
\usepackage{amssymb}
\usepackage{physics}
\usepackage{fancyhdr}
\fancypagestyle{alim}{\fancyhf{}\fancyfoot[R]{978-1-7281-4352-1/19/\$31.00 \copyright2019 IEEE}}
\title{Automatic Quantification of Facial Asymmetry using Facial Landmarks}
\twoauthors
 {Abu Md Niamul Taufique, Andreas Savakis}
	{Rochester Institute of Technology
	}
 {Jonathan Leckenby}
	{University of Rochester Medical Center
	}
\begin{document}
%
\maketitle
\begin{abstract}
One-sided facial paralysis causes uneven movements of facial muscles
on the sides of the face. Physicians currently assess facial asymmetry in a subjective manner based on their clinical experience. This paper proposes a novel method to provide an objective and quantitative asymmetry score for frontal faces. Our metric has the potential to help physicians for diagnosis as well as monitoring the rehabilitation of patients with one-sided facial paralysis. A deep learning based landmark detection technique is used to estimate style invariant facial landmark points and dense optical flow is used to generate motion maps from a short sequence of frames. Six face regions are considered corresponding to the left and right parts of the forehead, eyes, and mouth. Motion is computed and compared between the left and the right parts of each region of interest to estimate the symmetry score. For testing, asymmetric sequences are synthetically generated from a facial expression dataset. A score equation is developed to quantify symmetry in both symmetric and asymmetric face sequences. 
\end{abstract}

\section{Introduction}
Asymmetry in facial expression can be a sign of Bell's Palsy or the result of a stroke. 
Bell's Palsy causes immobilization of one side of the face for 99\% of the patients. Studies show that patients who suffer from severe Bell's Palsy also face psychosocial difficulty due to altered appearance \cite{hohman2014determining}.

Quantifying the asymmetry in facial expression can play a significant role during treatment of the Bell's Palsy patients. It can help the treatment procedure in two ways, first by providing surgical guideline and then for monitoring the recovery.
 
Physicians typically estimate facial asymmetry in a subjective manner based on their clinical experience. The widely used subjective grading system is the House-Brackmann grading system \cite{house1983facial}. 
Objective methods take a more rigorous approach to measure and quantify facial asymmetry.
They can enhance the understanding of asymmetry beyond the subjective grading system by keeping track of objective metrics during the course of rehabilitation. 
Several objective methods are proposed in the literature to estimate the degree of asymmetry in a facial expression \cite{bajaj1997quantitative, johnson1997quantitative, wang2004objective, wang2016automatic, he2007biomedical}.
Bajaj-Luthra et al. \cite{bajaj1997quantitative} uses anatomic or non-anatomic motion to quantify facial asymmetry. Johnson et al. \cite{johnson1997quantitative} used a similar method to compute the muscle transfer of paralysed patients. 

In this paper, our contribution is two fold. We leverage state-of-the-art landmark detection to localize facial landmarks, which form the basis of splitting the face into chosen regions. 
We then utilize dense optical flow estimation and develop a novel way of estimating the facial symmetry score from the flow magnitude in symmetric regions of the face.
\begin{figure*}[ht]
\centerline{\includegraphics[width=0.9\textwidth]{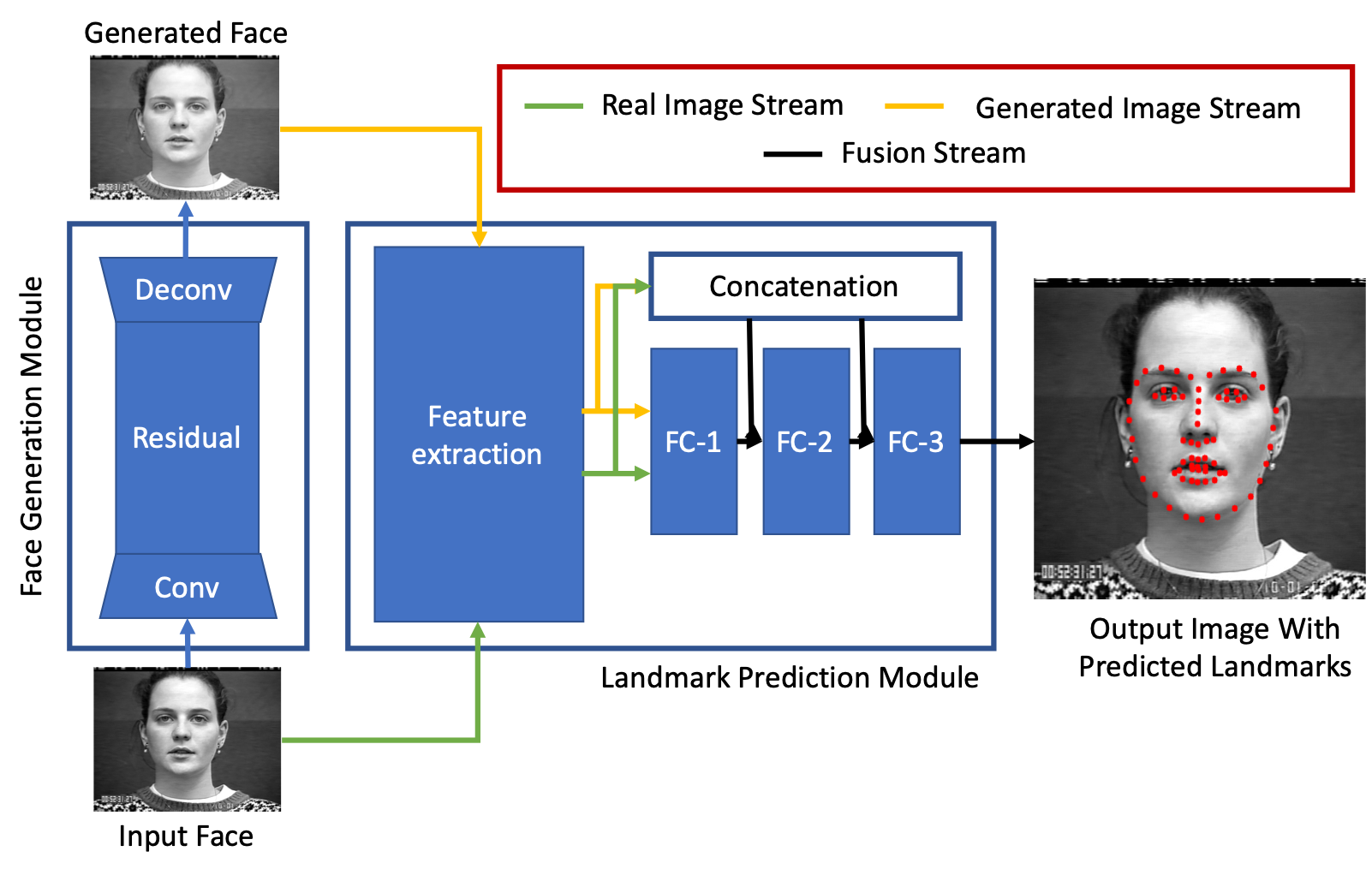}}
\caption{Style Aggregated Network (SAN) method for facial landmark detection \cite{dong2018style}. Best viewed in color.}
\label{fig:network}
\end{figure*}

\begin{figure}[t]
\centerline{\includegraphics[width = 0.5\textwidth]{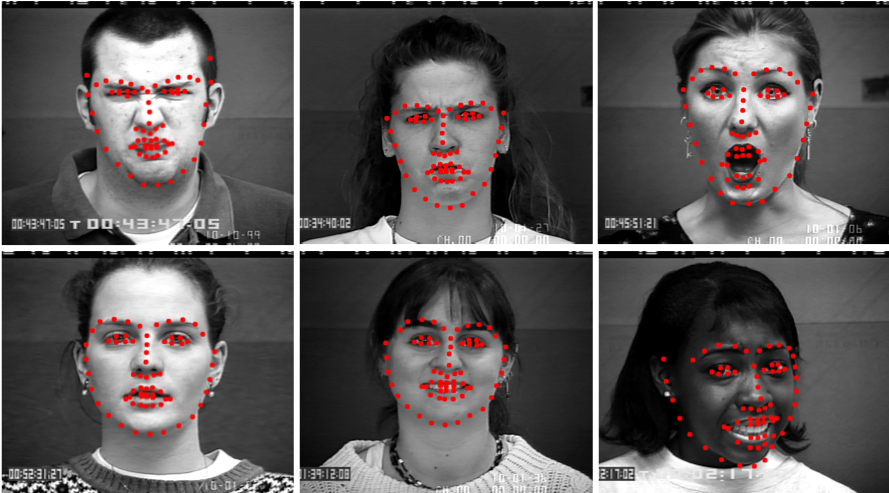}}
\caption{Examples of facial landmark localization using the SAN method on expressive faces.}
\label{fig:landmarks}
\end{figure}
\section{Related Work}
An early quantitative approach for facial asymmetry quantification is using $D_{face}$ technique proposed by Liu and Mitra et al. \cite{liu2003facial}. This technique is used for computing asymmetry from a single expressive image. $D_{face}$ is computed by extracting the density image from an expressive image and taking the difference between the left side and the right side of the face. 
Wang et at. proposed $P_{face}$ and eigenflow to measure the asymmetry between two sides of the face where $P_{face}$ uses $D_{face}$ on a sequence of frames \cite{wang2004objective}.
Another automatic method for estimating the facial asymmetry is proposed in \cite{wang2016automatic} where ASM is used for facial landmark detection and SVM is used for scoring facial asymmetry. 


The standard method that physicians use to perform facial symmetry assessment is eFACE \cite{banks2015clinician}, a clinician-graded scale that generates an overall facial disfigurement score. The eFACE method requires 16 parameters to be determined from a patient's face and performs multiple regression analysis to best fit expert-determined scores.  

In \cite{he2007biomedical}, optical flow is used for assessing the movements of different parts of the face to estimate asymmetry. Regions of interest are selected based on the edges found after edge detection. Then, the flow features are used in SVM to compute the asymmetry scores. However, edge finding algorithms may suffer from stability under variable illumination and may be sensitive to scale variation.  In our work, we utilize optical flow in regions determined using facial landmarks.

\section{Methodology}
We begin our processing pipeline with face detection using the Viola-Jones method \cite{viola2001rapid}.  
For facial asymmetry estimation we require a frontal face, which is well-suited for the Viola-Jones approach. 
We leverage the OpenCV implementation of this Haar feature-based face detection.

After face detection, we proceed with facial landmark detection to identify the important face regions, namely forehead, eyes and cheeks on both sides of the face. Then we compute the dense optical flow \cite{farneback2003two} in these regions and use a formula we developed to obtain a facial asymmetry score.

\begin{figure*}[tb]
\centerline{\includegraphics[width=\textwidth]{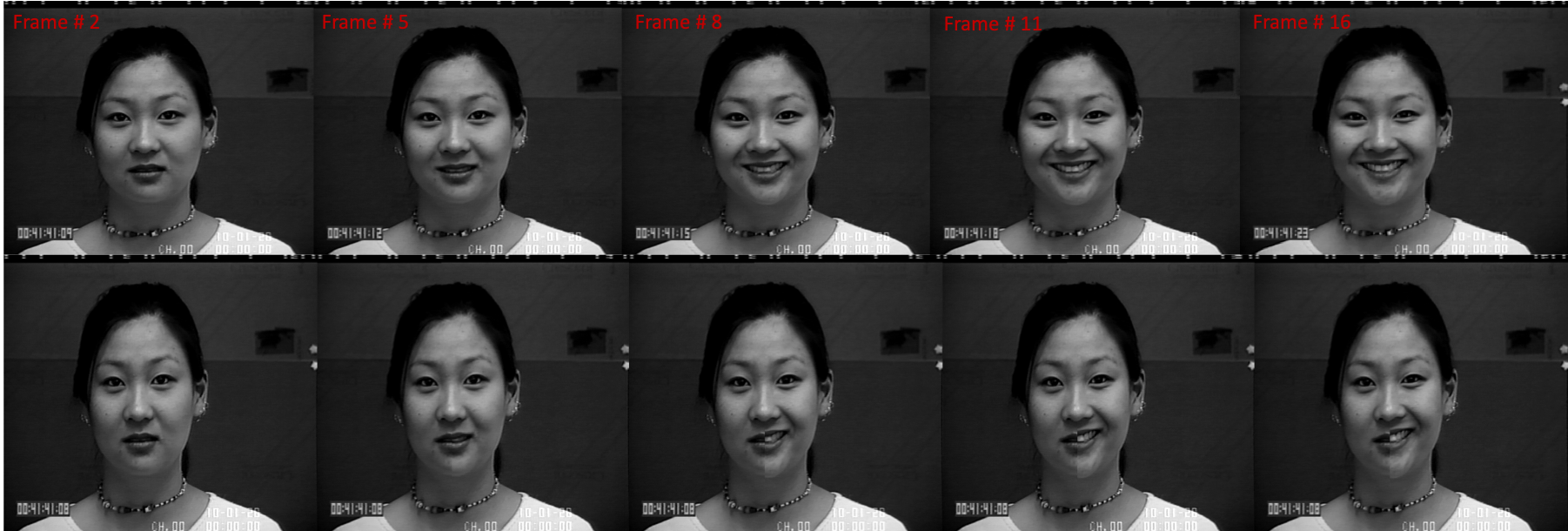}}
\caption{Creating an asymmetric sequence from a symmetric sequence.}
\label{fig:asymmetric-sequence}
\end{figure*}

\subsection{Landmark Detection}
Landmark detection is a fundamental step of our approach. We use the Style Aggregated Network (SAN) \cite{dong2018style} for facial landmark detection. SAN utilizes a style transfer technique to aggregate faces and extract discriminative features from them. 
Style aggregation helps achieve better discrimination, because during training it takes into account the intrinsic variation among different samples.

The SAN method is overviewed in Figure \ref{fig:network}.
CycleGAN \cite{zhu2017unpaired} is used to generate styled face from the original face. To train the face generation module, the authors first generated three styles (PS-generated) Light, Grey, and Sketch of various faces from the original dataset. A combined dataset of these styled faces and the original faces is used to fine-tune a ResNet-152 classifier with $C+1$ classes where $C=3$ is the number of styles. After fine-tuning, the original dataset (without the PS-generated images) is passed through the classifier and features are leveraged from the Global Average Pooling (GAP) layer. Then $k$-means clustering is performed on these features to find $k\!=\!3$ clusters of hidden styles within the original dataset. Finally, images of different styles are used to train the CycleGAN to generate styled faces. The generated faces are aggregated with the original faces for the subsequent steps of SAN. 

In the second phase of processing, the original and generated faces are fused using a feed forward network in a cascaded manner, as shown in Figure \ref{fig:network}. 
For this, features are extracted from the first four convolution blocks of VGG-16 \cite{simonyan2014very} followed by two additional feature extraction layers. The feature extraction stage takes an original face $\mathbf{I_o} \in \mathbb{R} ^{h\times w}$ and a generated face $\mathbf{I_g} \in \mathbb{R} ^{h\times w}$ as inputs and produces the output $\mathbf{F} \in \mathbb{R} ^{C \times {h'} \times {w'}}$ where, $C$ is the number of channels of the last convolution layer and $(h',w') = (h/8,w/8)$. The extracted features are denoted by $\mathbf{F_o}$ and $\mathbf{F_g}$ for the original and generated faces, respectively. The first fully convolutional stage (FC-1) takes the features of the faces, $\mathbf{F_o}$ and $\mathbf{F_g}$ as inputs and generates belief maps \cite{wei2016convolutional}, $\mathbf{B_o}$ and $\mathbf{B_f}$ for the original and generated faces, respectively. Next, the second fully convolutional stage (FC-2) takes the concatenation of $\mathbf{F_o}$, $\mathbf{F_g}$, $\mathbf{B_o}$, and $\mathbf{B_f}$ as input and predicts a belief map $\mathbf{B_2}$. 
\begin{equation}
s2(\mathbf{F_o}, \mathbf{F_g}, \mathbf{B_o}, \mathbf{B_g}) = \mathbf{B_2}
\end{equation}
The third and final fully convolutional stage (FC-3) takes the concatenation of the output from stage 2, $\mathbf{B_2}$ along with the facial features, $\mathbf{F_o}$ and $\mathbf{F_g}$ as input, and predicts the final belief map $\mathbf{B_3}$, which is used to predict the coordinates of the landmarks.         
\begin{equation}
s3(\mathbf{F_o}, \mathbf{F_g}, \mathbf{B_2}) = \mathbf{B_3}
\end{equation}
The loss function $L$ compares the belief maps $\mathbf{B_i}$ and corresponding ideal maps  $\mathbf{B_i^*}$
as follows,
\begin{equation}
L = \sum_{i \in \{ o,g,2,3\}} \left\Vert \mathbf{B_i - B_i^*} \right\Vert_F ^2 .
\end{equation}

The landmark locations of some expressive images form the extended Cohn-Kanade dataset \cite{lucey2010extended} are shown in Figure \ref{fig:landmarks}. 
Based on the landmark locations, we select the regions corresponding to the forehead, eyes, and cheeks (including mouth). After extracting these main face regions, we split each region into left and right parts symmetrically, for instance, left and right eye regions. Then we compare the total optical flow between the left and the right parts to compute the symmetry score.

\subsection{Symmetry Score}
To get an objective score, we compute the total movement per pixel per frame for each face region in the video sequence. If $f(x,y,t)$ is the flow magnitude computed from $t^{th}$ and $(t-1)^{th}$ frames where $x$ and $y$ represent the pixel space, the  Movement Score $V_S$, over all frames, can be written as 
\begin{equation}
V_S =\frac{1}{M (N-1)}  \sum\limits_{t = 2}^{N} f(x,y,t)
\end{equation}
where $M$ is the total number of pixels within the region of interest and $N$ is the total number of frames. From this we can compute the Symmetry Score $S_S$ between the left and right sides of a region as:
\begin{equation}
    S_S = 1 - \lambda\abs{V_{S_{left}} - V_{S_{right}}}
\end{equation}
where $\lambda$ is a coefficient determined from ground truth asymmetry scores obtained by the evaluation of experts. 
We found $\lambda = 3.8$ in our study. 
If the Symmetry Score goes above $1$ or below $0$, it is set to $1$ or $0$, respectively. 

\begin{figure*}[ht]
\centerline{\includegraphics[width=\textwidth]{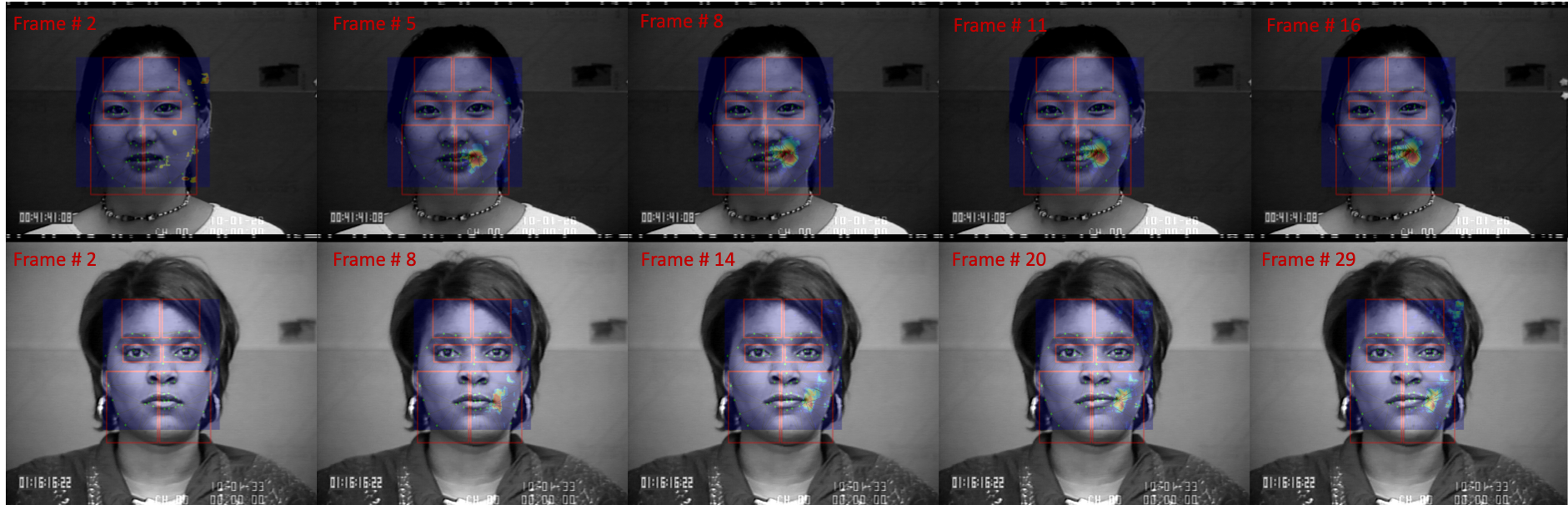}}
\caption{Evaluation of asymmetric sequences. The top row, middle row, and the bottom row represent sequence 128-011 and 133-001, respectively. At first, these sequences are converted to asymmetric sequences and then total flow is estimated. Best viewed in color.}
\label{fig:asymmetric}
\end{figure*}

\section{Experiments and Results}
In this section, we present the dataset and experimental set-up of our approach.
Due to lack of publicly available datasets with asymmetric faces, we generated our own asymmetric sequences.

\subsection{Asymmetric Face Sequences}

To obtain data, we created asymmetric sequences from the symmetric sequences of the CK+ dataset \cite{lucey2010extended}. Landmark face locations are used to split the face into two halves, left and right. We kept the left side of the face unaltered and fixed the right side as it appears in the first frame of the sequence. With this simple yet effective technique, we were able to create asymmetric face sequences. The creation of an asymmetric sequence is shown in Figure \ref{fig:asymmetric-sequence}.

\subsection{Experimental Setup}
\label{sec:expsetup}
We select the regions of interest within the face based on the landmark points computed by the SAN method.  
In the first frame, we extract the face patch using the face detection algorithm. For the rest of the frames we compute the magnitude of the optical flow and sum them up pixel wise temporally to get the total movement within the facial region. 

The total optical flow between the left and the right part of each region is compared and the symmetry score is computed. 
For SAN, we have used the pretrained network available from their Github repository. The network is pretrained on 300-W \cite{sagonas2013300} dataset.
 
\subsection{Results}




Since the optical flow estimation can be noisy in flat regions, we use thresholding to reduce noise. 
We used a frame by frame thresholding where any flow magnitude less than six times the average magnitude within the cropped region is set to zero. 
The flow magnitude heatmaps overlayed with the face images of two asymmetric sequences are shown in Figure \ref{fig:asymmetric}.



We have asked one expert to provide scores for the original and the generated sequences for the sequences $128-011$ and $133-011$,  
shown in Tables \ref{tab:symmetryscores} and \ref{tab:asymmetryscores}, respectively. 
The tables show the results of the measured symmetry scores using our algorithm and the expert opinion and indicate that there is close agreement. 

\begin{table}[ht]
\caption{Symmetry Scores for the original symmetric sequences. The symmetry score obtained with our method is denoted as $S_S$ and the expert opinion is denoted as $E_O$.} 
\label{tab:symmetryscores}
\begin{center}        
\begin{tabular}{c|cc|cc|cc} 
\hline\hline
Seq & \multicolumn{2}{c|}{Eye} & \multicolumn{2}{c|}{Forehead} & \multicolumn{2}{c}{Cheek} \\
 & $S_S$ & $E_O$ & $S_S$ & $E_O$ & $S_S$ & $E_O$ \\
\hline\hline
S128-011 & 1 & 1 & 1 & 1 & 0.9 & 1 \\
S133-001 & 1 & 1 & 1 & 1 & 1 & 1 \\
\hline\hline
\end{tabular}
\end{center}

\caption{Symmetry Scores for the asymmetric sequences. The symmetry score obtained with our method is denoted as $S_S$ and the expert opinion is denoted as $E_O$.}
\label{tab:asymmetryscores}
\begin{center}        
\begin{tabular}{c|cc|cc|cc} 
\hline\hline
Seq & \multicolumn{2}{c|}{Eye} & \multicolumn{2}{c|}{Forehead} & \multicolumn{2}{c}{Cheek} \\
 & $S_S$ & $E_O$ & $S_S$ & $E_O$ & $S_S$ & $E_O$ \\
\hline\hline
S128-011 & 1 & 1 & 1 & 1 & 0.58 & 0.6  \\ 
S133-001 & 0.97 & 1 & 0.99 & 1 & 0.92 & 0.9 \\
\hline\hline
\end{tabular}
\end{center}
\end{table}

\section{Conclusions}
We developed a novel technique to automatically quantify  facial asymmetry in videos for patient treatment and rehabilitation. A state-of-the-art landmark detection technique is used for style invariant landmark detection and region of interest are generated using the landmark locations. Optical flow is used to compute the motion in six facial regions of interest, and a symmetry score is obtained based on the motion aggregation. In future work, we would like to work with a real dataset of facial asymmetry and create an end-to-end trainable facial asymmetry method.

\bibliographystyle{IEEEbib}
\bibliography{refs}

\end{document}